\documentclass[11pt]{ouparticle}  

\usepackage{makecell}
\usepackage{multirow}
\usepackage{booktabs}
\usepackage{upgreek}
\usepackage{array}
\usepackage{amsmath}
\usepackage{amssymb}
\usepackage{bm}
\usepackage{cite}
\usepackage{subfigure}
\usepackage{hyperref} 
\hypersetup{
    bookmarks=true,         
    unicode=false,          
    pdftoolbar=true,        
    pdfmenubar=true,        
    pdffitwindow=false,     
    pdfstartview={FitH},    
    pdftitle={NSR},    
    pdfauthor={Camps-Valls},     
    pdfsubject={NSR},   
    pdfcreator={Camps-Valls},   
    pdfproducer={}, 
    pdfkeywords={}{}{}{}{}, 
    pdfnewwindow=true,      
    colorlinks=true,       
    linkcolor={black},          
    citecolor=gray,        
    filecolor=gray,      
    urlcolor=gray           
}

\usepackage{graphicx} 
\usepackage{graphics}

\usepackage{color}
\usepackage{colortbl}

\definecolor{MYCOLOR0}{rgb}{0.92,0.92,0.92}
\definecolor{MYCOLOR}{rgb}{1,1,0}
\definecolor{MYCOLOR2}{rgb}{0.5,1,0.5}
\definecolor{MYCOLOR3}{rgb}{0.88,1,1}

\def\x{{\mathbf x}}

\def\f{{\mathbf f}}
\def\g{{\mathbf g}}
\def\y{{\mathbf y}}

\newcommand{\dataset}{{\cal D}}

\usepackage{pgf}
\usepackage{tikz}
\usetikzlibrary{arrows,automata,positioning}
\usetikzlibrary{shapes,arrows,positioning}
\tikzset{
    >=stealth',
    punkt/.style={
           rectangle,
           rounded corners,
           draw=black, very thick,
           text width=6.5em,
           minimum height=2em,
           text centered},
    pil/.style={
           ->,
           thick,
           shorten <=2pt,
           shorten >=2pt,}
}

\begin{document}

\title{A Perspective on Gaussian Processes for Earth Observation}

\author{%
\name{Gustau Camps-Valls$^1$, Dino Sejdinovic$^2$, Jakob Runge$^3$, Markus Reichstein$^4$}
\thanks{Published in National Science Review 6 (4):616-618, 2019 DOI: https://academic.oup.com/nsr/article/6/4/616/5369430.}
\address{
$^1$Image Processing Laboratory, Universitat de Val\`encia, Val\`encia, Spain\\
$^2$Department of Statistics, University of Oxford, Oxford, UK\\
$^3$German Aerospace Center, Institute of Data Science, Jena, Germany\\
$^4$Max Planck Institute for Biogeochemistry, Jena, Germany}
}

\date{}

\maketitle

\vspace{-2cm}
\section*{Introduction}
\label{sec1}

Earth observation (EO) by airborne and satellite remote sensing and in-situ observations play a fundamental role in monitoring our planet. In the last decade, machine learning has attained outstanding results in the estimation of bio-geo-physical variables from the acquired images at local and global scales in a time-resolved manner. 
Gaussian processes (GPs)~\cite{Rasmussen06}, as flexible nonparametric models to find functional relationships, have excelled in EO problems in recent years, mainly introduced for model inversion and emulation of complex codes~\cite{CampsValls16grsm}.
GPs provide not only accurate estimates but also principled uncertainty estimates for the predictions. Besides, GPs can easily accommodate multimodal data coming from different sensors and from multitemporal acquisitions. 
Due to their solid Bayesian formalism, GPs can include prior physical knowledge about the problem, and allow for a formal treatment of uncertainty quantification and error propagation. 

\begin{figure}[h!]
\centerline{\includegraphics[width=7.5cm]{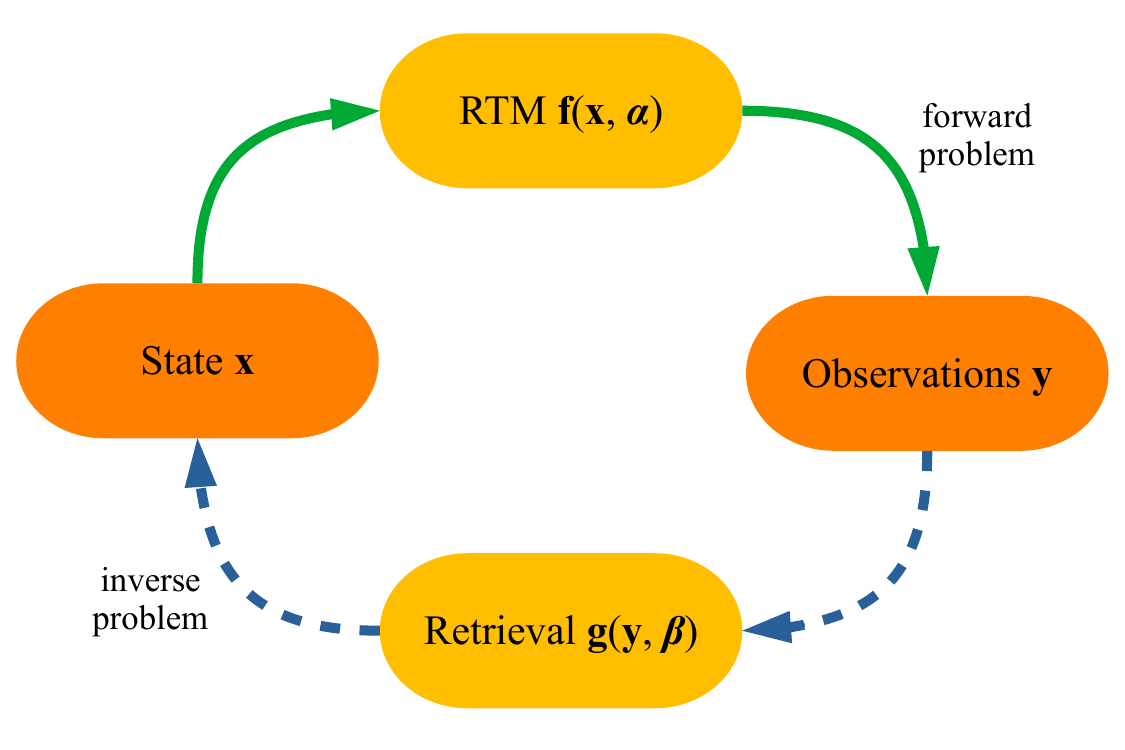}}
\vspace{-0.25cm}
\caption{Forward (solid lines) and inverse (dashed lines) modelling in Earth observation.} 
\label{forward_inverse}
\end{figure}

In remote sensing, we often deal with {\em radiative transfer models} (RTMs) which implement the equations of energy transfer. These codes are needed for modelling, understanding, and predicting some variables of interest related to the state of the land cover, water bodies and atmosphere.
An RTM $\f$ operating in {\em forward mode} generates a multidimensional {\em radiance observation} $\y\in\mathbb{R}^p$ seen by the sensor given a multidimensional {\em parameter state vector} $\x\in\mathbb{R}^d$, see Fig.~\ref{forward_inverse}. 
Running forward simulations yields a look-up-table (LUT) of input-output pairs, $\dataset=\{(\x_i,\y_i)\}_{i=1}^n$. Solving the {\em inverse problem} implies learning the function $\g$ using $\dataset$ to return an estimate $\x_*$ each time a new satellite observation $\y_*$ is acquired. 
GPs have been used to learn both the often costly forward model $\f$ as well as the inverse model $\g$. Learning the forward model allows for faster simulations, while learning an inverse model has allowed to provide physically-meaningful, spatially-explicit, and temporally-resolved maps of variables of interest.

Despite great advances in forward and inverse modelling, GP models still have to face important challenges, such as the high computational cost involved or the derivation of faithful confidence intervals. 
More importantly, we posit that GP models should evolve towards {\em data-driven physics-aware models} that respect signal characteristics, be consistent with elementary laws of physics, and move from pure regression to observational {\em causal inference}. 

\newpage

\section*{Advances in GP inverse modelling}

The most important shortcoming of GPs is their high computational cost and the memory requirements, which grows cubically and quadratically with the number of training points, respectively. 
Recently, a great progress has been made in constructing scalable versions of GPs, demonstrating their utility in big data regimes~\cite{hensman2013gaussian}. 

An important challenge in Earth observation relates to the fact that data comes with complex nonlinearities, levels and sources of noise, and non-stationarities. Standard GPs often assume homoscedastic noise and use stationary kernels though. 
The current state-of-the-art GP to deal with heteroscedastic noise makes use of 
a marginalized variational approximation~\cite{CampsValls16grsm}. 
The method has resulted in excellent performance in estimating biophysical parameters (chlorophyll-a content in plants and water bodies) from acquired reflectances. 
In many EO applications one transforms the observed variable to linearize or Gaussianize the data via parametric transforms. A {\em warped} GP model has allowed learning a non-parametric optimal transformation from data, and has shown very good results in predicting vegetation parameters (chlorophyll, leaf area index, and fractional vegetation cover) from hyperspectral images~\cite{Mateo18wgp}. 
Another common problem in remote sensing is that of ensuring consistency across products: estimating several related variables simultaneously can incorporate their relations in a single model. 
A recent latent force model (LFM) GP can encode ordinary/partial differential equations governing the system, and has allowed to monitoring crops, estimate multiple vegetation covariates simultaneously, and deal with missing observations due to the presence of clouds or sensor acquisition problems~\cite{CampsValls18sciasi}. 

Making inferences with GPs is not only about obtaining point-wise estimates but also faithful uncertainty estimates, essential to perform error propagation. Inference should also contemplate {\em extrapolation} analysis as an ambitious far-end goal. 
Besides, note that we ultimately aim to characterize model error by comparing simulators to reality, calibrate models by proper estimation of (hyper)parameters, and make uncertainty statements about the world that combine models, data, and their corresponding errors. We think that the Bayesian formalism of GPs is the natural framework to tackle these yet unresolved problems.

\section*{Advances in GP forward modelling}

Surrogate modelling, also known as {\em emulation}, based on GPs is gaining popularity in remote sensing. 
Emulators are essentially statistical models that learn to mimic the RTM code using a representative dataset $\dataset$. GPs have largely dominated the field for decades and have provided excellent accuracy and physical consistency as studied via sensitivity analysis in the context of vegetation and atmosphere models in~\cite{CampsValls16grsm}. Once the GP model is trained, one can readily perform fast forward simulations, which in turn allows improved inversion. However, replacing an RTM with a GP model requires running expensive evaluations of $\f$ first. Recent more efficient alternatives construct an approximation to $\f$ starting with a set of support points selected iteratively~\cite{CampsValls18sciasi}. 
This topic is related to active learning and Bayesian optimization, which might push results further in accuracy and sparsity, especially when modelling complex codes. 

RTMs are the result of many decades of scientific research and continuous development, so they often include {\em ad hoc} rules, heuristics, and non-differentiable links that hamper analytic treatment. Emulation allows to account for input errors, derive predictive variance estimates, infer sensitivity values of parameters, calculate Jacobians, and perform uncertainty propagation and quantification analytically. 
Besides, a lot of physical knowledge used for designing RTMs could be translated in designing priors (e.g. physically plausible parameter values). 
These excellent capabilities have not been widely exploited in EO applications though. 


\section*{Towards physics-aware GP modelling}
\label{sec3}

The GP framework allows us to include constraints and priors adapted to {\em signal features} such as non-stationarity, circularity, spatial-temporal relations, coloured-noise processes, and non-i.i.d. relations. Nevertheless, data-driven GP models should be further constrained to provide physically-plausible predictions. 
Recent approaches consider designing joint observation-simulation cross-covariances~\cite{CampsValls18sciasi}. 
Recently we suggested a full framework for hybrid modelling with machine learning~\cite{reichstein18nat}, which could be formalized within the GP probabilistic framework too. 

Learning dynamical physical systems is very challenging. Recent regression approaches have learned the governing equations of nonlinear dynamical systems from data, such as the Lorenz, Navier-Stokes and Schr\"odinger equations. Models typically impose sparsity and hierarchical modelling, but also a GP probabilistic approach has excelled in discovering ordinary and partial differential, integro-differential, and fractional order operators~\cite{Raissi17}. 

The integration of physics into GP models does not only achieve improved generalization but, more importantly, endorses these grey-box models with {\em consistency} and {\em faithfulness}. As a by-product, the hybridization process has an interesting regularization effect, as physics discards implausible models and promotes simpler structures. 

\section*{From regression to causation}

Understanding is more challenging than predicting, especially when no interventional studies can be conducted, as in the Earth sciences. Causal inference from observational data to estimate causal graphical models has  become a mature science with effective machine learning methods to deal with  both time series and non-time ordered data, see \cite{Peters18,ZhaSchSpiGly17} and references therein. Causal inference methods can be classified roughly into conditional independence or constraint-based approaches and structural causal models. Constraint-based causal discovery algorithms iteratively infer graphical models utilizing conditional independence testing. In \cite{Runge2018d} a GP-based conditional independence test is combined with a scalable causal discovery algorithm allowing to infer high-dimensional graphical models from time series data. Constraint-based algorithms only allow to infer causal graphical models up to a Markov equivalence class. Utilizing additional assumptions, such as on the noise distribution or functional dependence, the class of structural causal models \cite{Peters18} allows to infer causal directionality in such undecidable Markov equivalent cases. Further GP-based causal discovery methods include \cite{HuaZhaSch15} where a GP model was used as a prior to capture the time-varying causal association in a non-parametric manner, while in \cite{FlaxmanNS16} GPs were exploited as an efficient pre-whitening step to deal with non-iid observations so common in remote sensing. 
Recently, \cite{Mateo18wgp,CampsValls16grsm} introduced the WGP regression in additive noise models to account for post-nonlinear effects and heteroscedastic noise respectively, and applied it successfully to a set of geoscience and remote sensing bivariate problems. Some important challenges in causal inference for the Earth science are still to be solved: how to scale GP models to deal with millions of points, missing data and time aggregation as well as time sub-sampling, and complex spatial-temporal dependency structures. Testing scientific hypotheses, comparing model-vs-data causal graphs, and assessing the impacts of extreme events, are just some exciting further avenues of research. 


\section*{Funding}

GCV would like to acknowledge the support from the European Research Council (ERC) under the ERC Consolidator Grant 2014 project SEDAL (grant agreement 647423). 

\small
\bibliographystyle{unsrt} 
\bibliography{nsr}

\end{document}